# Advancing Wildlife Monitoring: Drone-Based Sampling for Roe Deer Density Estimation

Stephanie Wohlfahrt[1], Christoph Praschl[2], Horst Leitner[1], Wolfram Jantsch[1], Julia Konic[3], Silvio Schueler[3], Andreas Stöckl[2] and David C. Schedl[2],

[1]*Office for Wildlife Ecology and Forestry, Klagenfurt, Austria*

[2]*Digital Media Lab, University of Applied Sciences Upper Austria, Hagenberg, Austria*

[3]*Austrian Research Centre for Forests BFW, Vienna, Austria*

04.08.2025, Klagenfurt am Wörthersee, Austria

**Abstract**

Our study uses unmanned aerial drones to estimate wildlife density in southeastern Austria. We compare our drone-based estimates to camera trap data and identify apparent differences.

Accurate wildlife density estimates are crucial for informed wildlife management. Traditional methods for estimating density, such as capture-recapture, distance sampling, or camera traps, are well-established but labour-intensive or spatially constrained. Using thermal signatures (IR) and RGB imagery, unmanned aerial drones enable efficient, non-intrusive animal counting from the air. Our surveys were conducted during the leafless period on single days in October and November 2024 in three survey areas within a sub-Illyrian hill and terrace landscape. The flight transects were determined based on predefined drone launch points, using a 350 m grid and an algorithm to define the flight direction of consecutive systematically randomized transects. This allows large areas to be surveyed in a single day, simultaneously with multiple drones, and minimizes double counts of animals. The flight altitude was set at 60 m to avoid disturbing the target species, roe deer (Capreolus capreolus), while ensuring animal detection.

The number and species of animals per transect were manually annotated in the recorded imagery and extrapolated to densities per 1 km$^2$ for the entire survey region. We applied three extrapolation methods with increasing levels of complexity: naïve area-based extrapolation, bootstrapping, and modelling using a zero-inflated negative binomial distribution. As a reference, we calculated a Random Encounter Model (REM) density estimate based on camera trap data from each area during the period of the drone flights. The three drone-based extrapolation methods yielded similar results, consistently showing higher density estimates than those derived from REM, except in one area in October.

We hypothesize that while the drone-derived density reflects wildlife density in forested and non-forested areas during the daytime on single days, REM density represents an averaged value over a month-long period, capturing activity throughout the entire day within forested areas. Although both methods aim to estimate density, they offer fundamentally different perspectives on wildlife activity.

Using drones for sample-based surveys presents a promising approach for density estimation.

**Introduction**

Modern, data-driven wildlife management is based on reliable estimates of the distribution and density of a population in a defined study area. The use of imagery, for example, from camera traps (CT) or unmanned drones, is already well established for this purpose. Drones can autonomously collect data along pre-programmed transects and at altitudes that do not disturb wildlife, which can then be analysed manually or using AI. For example, drones are being used to monitor crocodile nests from the air (Evans et al. 2016), estimate jellyfish densities (Hamel et al. 2021) or count ungulates (Ito et al. 2022). The accuracy and reliability of thermal drones for ungulate density estimations have already been tested in a study of white-tailed deer (*Odocoileus virginianus*) (Beaver et al. 2020).

Flight transects are used to randomly sample sub-areas of the survey area. This allows large areas to be surveyed in a single day while conserving resources. In studies of ungulates, parallel transects were chosen to cover the entire area to avoid potential double-counting of moving animals (Beaver et al. 2020, Baldwin et al. 2023). However, the size of the study area and the shape of the terrain may limit feasibility. An alternative would be to select a number of shorter transects systematically randomly distributed across the area in a multidirectional manner.

Depending on the individual detectability of the target species, CTs are used for capture-recapture methods (Seber 1974), distance sampling (Howe et al. 2017) or the random encounter model (REM) (Rowcliffe et al. 2008), which is particularly suitable for indistinguishable species such as roe deer (*Capreolus capreolus*). A minimum of 40, and ideally 100, measurable encounters of the target species with the CTs should be recorded to achieve sufficient accuracy in REM density estimation (Palencia et al. 2024). To achieve this, CTs need to be active for a minimum period, which makes it difficult to sample agricultural areas, for example. One solution is to limit the survey area to forest areas. As roe deer use both forest and non-forest areas as habitat (Lorenzini et al. 2022), CTs placed in this way can only provide an average estimate of roe deer density in forest areas and not over the entire study area.

In our study, we aim to test the feasibility of using drones along systematically randomized transects for density estimation and compare the results with those of CTs in forest areas. To exclude double counting and to obtain a systematically randomized transect selection that can be surveyed in one day, we used a simple algorithm based on possible drone starting points. Our hypothesis is that the two methods measure densities at different scales, as the CTs provide an averaged value over the duration of one month and over the entire 24h day in forest areas, while the drone method provides an insight into the density on one day, during the day and in forest and non-forest areas.

**Methods**

The three study areas are in south-eastern Austria in a sub-Illyrian hill and terrace landscape and extend between 47.03281335 degrees north latitude and 46.80255756 degrees south latitude, as well as 15.98127187 degrees east longitude and 15.71521058 degrees west longitude. The areas range in size from 2.98 to 5.49 km$^2$ and in altitude from 267 to 476 metres above sea level. The submontane level is dominated by beech forests (*Fagus sylvaticus*), often mixed with fir (*Abies alba*), chestnut (*Castanea sativa*) and red pine (*Pinus sylvestris*). The region has a high level of ecological diversity, strongly influenced by soil and climatic conditions. The dominant ungulate species is roe deer, which is fed with roughage in winter in all three areas. The average annual rainfall is between 700 and 1250 mm (Kilian et al. 1994).

All three areas were flown by DJI M3T and DJI M30T drones during daylight hours on one day in late October and one day in mid-November. A selection algorithm was used to select flights of up to 7 connected transects, 350 m in length, based on possible starting points. At each turnpoint, the algorithm decided which direction the next transect should be flown according to a set of rules. This process was repeated until a maximum flight distance was reached. In this way, 28 to 47 transects per area and flight day were flown in several flights. The flight altitude was set at 60 m AGL to avoid disturbances and to maximise recording quality. The drones recorded RGB and thermal video sequences during the overflights, as well as the camera's extrinsics, including the global position and orientation. Video footage was analysed separately by two observers, and sightings were geo-referenced and summarized per transect.

Based on these data, three extrapolations of increasing complexity were made: i) a naive extrapolation from the total number of sightings in relation to the total area flown, ii) a bootstrapping procedure, and iii) modelling density using a zero-inflated negative binomial distribution. For the bootstrap extrapolation, a density was calculated as the number of individuals per km$^2$ from the summarized sightings and area flown per transect. Mean and 95% CI were derived from 1000 bootstrapping iterations. The short transect length of 350 m compared to other studies results in a significant proportion of transects without sightings (between 62.2 and 77.5%). We attempted to account for this by modelling the count data using the R package glmmTMB while controlling for transect size in a two-stage process that included detection probability in addition to negative binomial counting for over-dispersed data. The summation of posterior predicted values results in a density per km$^2$ with error propagated SEs (Buckland et al. 2001).

A density estimate derived from REM in the same period serves as a control in this study. For this purpose, 21 (Survey A) and 22 (Survey B + C) Patriot cameras (Browning International S.A., 84050 Morgan, Utah) were set up in a 350 m grid in forest areas. The cameras were mounted at the intersections of the grid, north-facing on trees at a height of approximately 50 cm, and were programmed to take eight consecutive photos without interruption when triggered. Each deer encounter with a camera was defined as a sequence as long as one or more animals crossed the field of view. Annotation of the sequences and extraction of the necessary REM parameters were performed in the Agouti[1] platform. Data analyses were performed in R software in RStudio/2024.12.1+563[2]. We used an analysis of variance (ANOVA) with a type II sum of squares to test for significant differences between flight estimates and CT densities.

## Results

In a total of 227 transects, between 11.9 and 25.5% of the area was covered per flight day and area (cf. Table 1). The number of roe deer sightings ranged from 21 to 37. With the exception of the October flight in area A, the three extrapolation methods are closer to each other and higher than the CT densities (see Figure 1). While the latter ranged from 13.4 (survey C) to 32.0 deer/km$^2$ (survey A), the drone estimates per flight day ranged from 27.0 to 64.3 deer/km$^2$ (both survey A). An analysis of variance shows weakly significant differences between the methods used (F = 3.57, p = 0.038). A post hoc Tukey test shows no differences between the three drone estimates in detail, but weakly significant differences between the bootstrapping and ZINB methods with CT density (p = 0.038 and p = 0.026, respectively).

Table 1: Flight transects and roe deer sightings in three survey areas for flights in October and November 2024.

| Area (size km$^2$) | Flight | Covered area (km$^2$) | Covered area (%) | Sightings | Transects with sightings | Total transects |
|---|---|---|---|---|---|---|
| Survey A | Oct | 0.76 | 25.5 | 21 | 9 | 40 |
| (2.98) | Nov | 0.51 | 17.1 | 25 | 8 | 28 |
| Survey B | Oct | 0.94 | 17.1 | 37 | 13 | 47 |
| (5.49) | Nov | 0.74 | 13.5 | 25 | 8 | 36 |
| Survey C | Oct | 0.93 | 17.4 | 35 | 17 | 45 |
| (5.36) | Nov | 0.64 | 11.9 | 23 | 11 | 31 |

---

[1] https://agouti.eu/
[2] www.rstudio.com

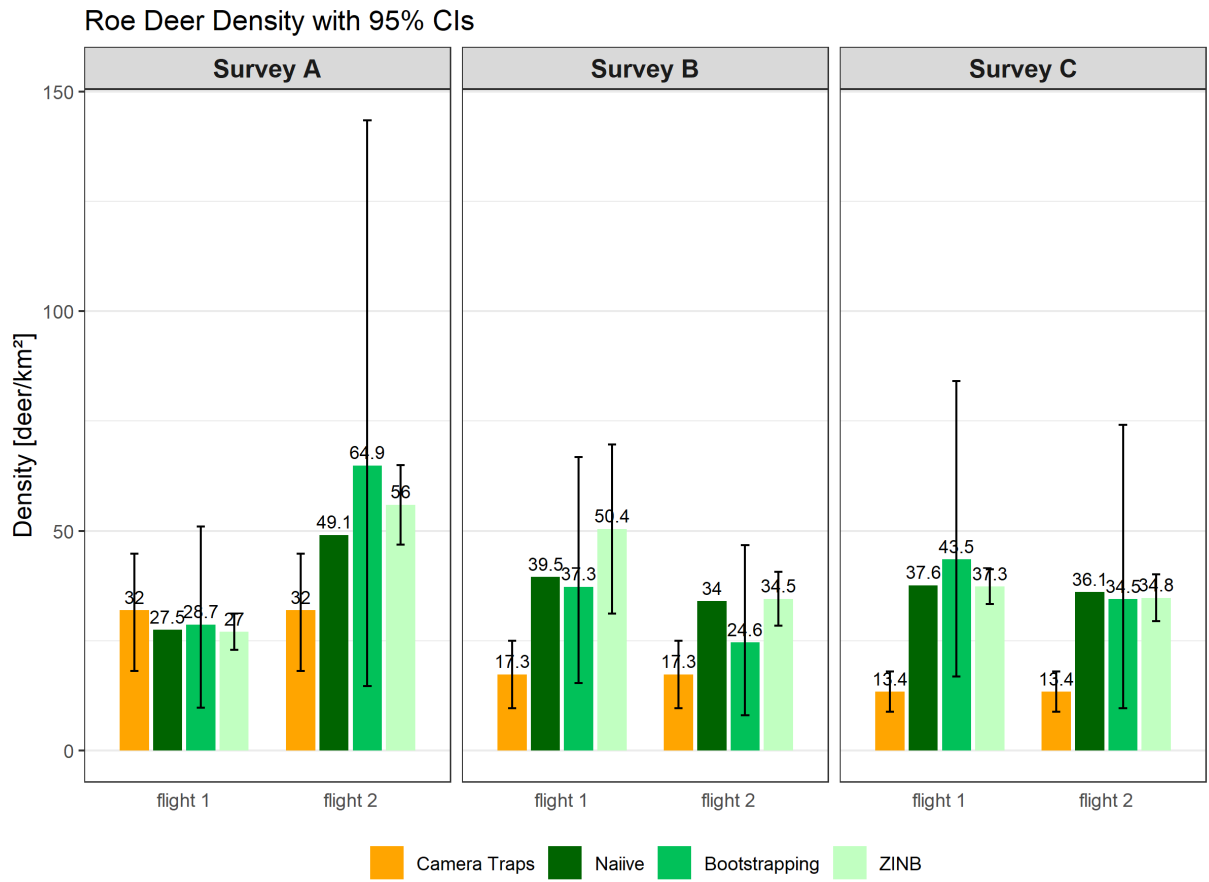

*Figure 1: Roe deer density estimations using CT data and REM (orange), and three extrapolation methods based on drone flight sightings (shades of green). Densities have been estimated for three survey areas (A, B, C) and two flight days in October (flight 1) and November 2024 (flight 2).*

**Discussion**

In our study, we were able to show that compared to a control roe deer density obtained using CT data and REM, drone densities were weakly significantly higher. Extrapolations at three levels of complexity resulted in similar density estimates without significant differences between them. This is a strong argument in favour of the simplest possible extrapolation using bootstrapping, which has been used in other studies (Beaver et al. 2020, Baldwin et al. 2023). In contrast to these studies, we calculated density per flight day and not as a mean of several flights. Considering the underlying distribution of the count data gave similar values to bootstrapping and naive extrapolation. Therefore, the naive method does not lead to biased values, but bootstrapping should be preferred to obtain 95% CIs.

We see the selection of systematically random transects covering all four cardinal directions as an advantage, as this allows a homogeneous sample to be collected, especially in larger and mountainous areas. Especially in these areas, transects exclusively along or orthogonal to the contour lines would potentially result in a biased sample, as the target species may also prefer to move along the contour lines. In a study in a fenced area of white-tailed deer with a known population size, density was calculated using parallel flown transects and extrapolation with a bootstrapping method to determine the 95% CIs (Beaver 2020). However, with an average density of 69.8 deer/km$^2$, there were no transects without sightings. So, unlike here, count data were available without a zero-inflated part of the distribution.

In our opinion, the difference between bootstrapped and modelled extrapolations and the CT density is a result of different sampling units. While CT density averaged over at least a month and over the full 24 h of the day exclusively in forest areas, the drone surveys provide an insight into the entire area exclusively during the day and on only one day. In Baldwin et al. (2023), ungulate densities calculated from parallel drone transects using bootstrapping were compared with mark-resight densities and N-mixture model (NMM) densities from CT data and yielded similar values (drone 31.3, NMM 27.5-40.7, mark-resight 27.7). In contrast to our study, the mean density of five flights was calculated rather than the density of two individual flights, as in our study. In addition, only one hour was flown around sunrise or sunset. An increase in the number of flight days or flights could lead to an approximation of the density value to the REM density value. Whether an extension of flights to night hours or a restriction to twilight hours would have a similar effect could be an interesting subject for further investigation.